\documentclass[11pt]{article}

\usepackage[final]{acl}

\usepackage{times}
\usepackage{latexsym}

\usepackage[T1]{fontenc}

\usepackage[utf8]{inputenc}

\usepackage{microtype}

\usepackage{inconsolata}

\usepackage{graphicx}

\usepackage{amsmath}
\usepackage{multirow}
\usepackage{booktabs} 
\usepackage{authblk}
\title{Clusters are All You Need: Pre-Training the Tsetlin Machine with Semantic Clusters from Language Models for Interpretability}

\author[1]{Jiechao Gao}
\author[2]{Rohan Kumar Yadav}
\author[3]{Yuangang Li}
\author[1]{Yuandong Pan}
\author[1]{\\Jie Wang}
\author[4]{Ying Liu}
\author[1]{Michael Lepech}
\affil[1]{Stanford University}
\affil[2]{Independent Researcher}
\affil[3]{University of California, Irvine}
\affil[4]{University of the Chinese Academy of Sciences}
\affil[1]{\texttt{\{jiechao, ydpan, jiewang, mlepech\}@stanford.edu}}
\affil[ ]{\textsuperscript{2}\texttt{errohanydv@gmail.com}, \textsuperscript{3}\texttt{yuanganl@uci.edu}, \textsuperscript{4}\texttt{liuy@ucas.ac.cn}}

\begin{document}
\maketitle
\begin{abstract}
Pre-trained language models such as BERT achieve strong text classification performance but lack transparency, limiting their use in high-stakes settings. The Tsetlin Machine (TM) offers fully interpretable, clause-based reasoning but captures little semantic information, and prior attempts to bridge the two rely on static word embeddings that miss contextual meaning. We propose a semantic pre-training framework that transfers knowledge from a pre-trained language model into a TM without using embeddings. Text samples are grouped into semantically coherent clusters with K-means or Top2Vec, and the resulting cluster–sample pairs pre-train a non-negated TM with enhanced Type I feedback. The TM thereby learns interpretable semantic keywords that are fine-tuned on downstream tasks. Across five datasets, our method substantially outperforms vanilla and embedding-based TMs and reaches performance competitive with BERT while remaining interpretable.
\end{abstract}

\section{Introduction}

The Tsetlin Machine (TM) has recently gained attention for its transparent, clause-based learning and explainable decisions \cite{Granmo2018TheTM}, with promising results on document classification \cite{10.24963/ijcai.2023/378}, sentiment analysis \cite{Yadav2021HumanLevelIL}, topic classification \cite{yadav-etal-2021-enhancing}, and fake news detection \cite{bhattarai-etal-2022-explainable}. Its inherent interpretability makes it attractive in high-stakes domains such as legal and medical text analysis, where transparency is essential \cite{Saha2022InterpretableTC,Berge2018UsingTT}.

However, TM operates on Boolean inputs, making Bag-of-Words (BOW) the most practical encoding. While BOW suits TM's logical mechanism, it prevents the model from leveraging the semantic knowledge of modern pre-trained language models, so TM often trails neural models initialized with large-scale pre-training. Existing attempts add static word embeddings such as Word2Vec, GloVe, or FastText to BOW \cite{yadav-etal-2021-enhancing}, but these still miss contextual semantics, and no interpretable method yet integrates transformer-based models such as BERT into TM.

\begin{figure}[t]
\centering
\includegraphics[width=0.45\textwidth]{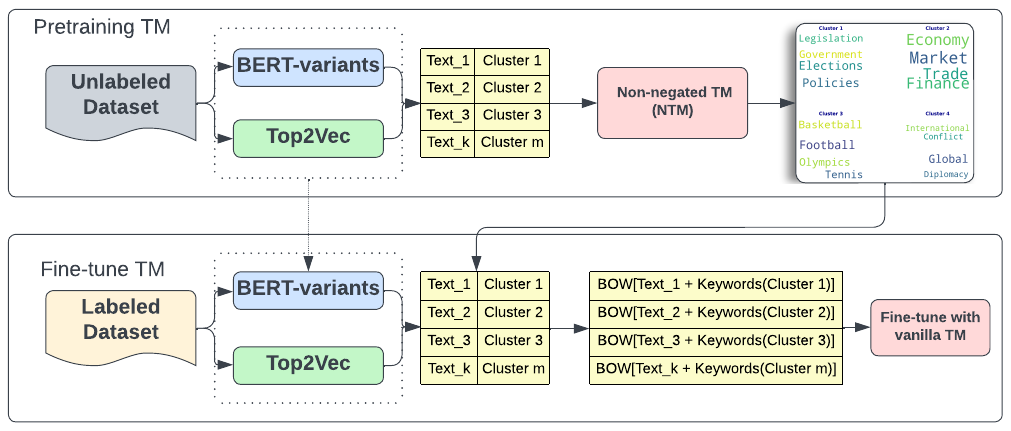}
\caption{Pre-training a Tsetlin Machine using semantic clusters from BERT or Top2Vec, followed by feature-extended fine-tuning with a vanilla TM.}
\label{fig1}
\end{figure}

In this paper, we propose a framework that transfers semantic knowledge from pre-trained language models into TM while preserving interpretability. As illustrated in Figure~\ref{fig1}, it follows two stages. First, unlabeled samples are embedded with a pre-trained language model and grouped into semantic clusters using either K-means or Top2Vec; the resulting cluster–sample pairs pre-train a Non-Negated Tsetlin Machine (NTM) that learns monotone, interpretable cluster representations without negated features, from which we extract high-confidence words or phrases as semantic descriptors for each cluster.

In the second stage, labeled samples are enriched with their cluster descriptors to form a semantic Bag-of-Words representation that fine-tunes a standard TM. Clustering and pre-training run once per domain and are reused across downstream tasks, amortizing their cost. Experiments on five datasets against nine baselines show consistent gains over vanilla TM and embedding-augmented TM, with performance comparable to BERT-based models, enabling effective and interpretable use of pre-trained semantic knowledge.

\section{Related Works}

Large-scale pre-trained language models such as BERT dominate NLP \cite{Devlin2019BERTPO,Raffel2019ExploringTL,10.5555/3495724.3495883}, learning general representations that are fine-tuned for downstream tasks and achieving state-of-the-art results on benchmarks such as GLUE \cite{Wang2018GLUEAM} and SuperGLUE \cite{10.5555/3454287.3454581}. However, these models remain opaque, which is problematic in sensitive domains where the decision process must be understood \cite{jain-wallace-2019-attention,Rudin2018StopEB,wu-etal-2023-transparency}. Transparent models \cite{molnar2022interpretable,10.24963/ijcai.2023/378} offer more insight but often sacrifice accuracy, whereas the Tsetlin Machine (TM) is notable for balancing the two. Integrating semantic embeddings into TM is difficult because of its Boolean BOW representation. A prior attempt encoding GloVe-similar words alongside BOW improved TM to match GloVe-initialized LSTM and CNN models \cite{yadav-etal-2021-enhancing}, but the field has since moved to richer pre-trained models such as BERT, which our work targets. An extended discussion is provided in Appendix~\ref{app:related}.


\section{Proposed Framework}

\begin{figure*}[!t]
\centering
\includegraphics[width=0.8\textwidth]{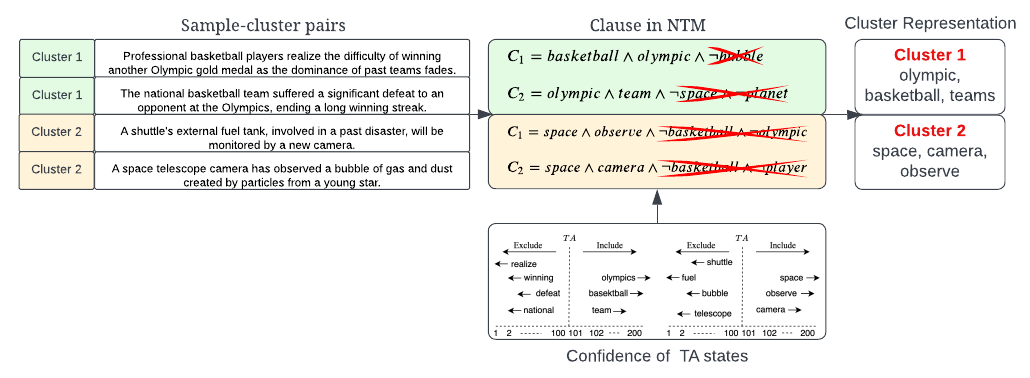}
\caption{Learning Process of pre-training NTM without negated feature and generating cluster representation with high confidence TA states.}
\label{fig2}
\end{figure*}
Here, we describe the proposed framework for pre-training the NTM, extracting words/phrases from the NTM, and then using these in downstream tasks with the vanilla TM. To better understand our framework, we give an overview of the basic TM architecture to understand how the vanilla TM functions in general in our supplemental materials.

In this section, we introduce the Non-Negated Tsetlin Machine (NTM), a variant designed to enhance interpretability and effectively utilize semantic information from pre-trained language models. NTM incorporates two critical modifications to improve its performance in capturing semantic relationships while maintaining transparency.

The primary enhancement in NTM involves boosting positive feedback (Type I feedback). Unlike the standard TM~\cite{Granmo2018TheTM}, where Type I feedback reinforces the inclusion of literals contributing to correct classifications, NTM significantly enhances this mechanism by directly incrementing the TA states associated with these literals. This adjustment prioritizes the reinforcement of TA states through positive feedback before considering other forms of feedback. By concentrating on strengthening features crucial for accurate predictions early in the learning process, NTM aims to consolidate learned patterns effectively. This refinement enhances the model's ability to capture subtle semantic relationships inherent in natural language data.
The Boosted True Positive TM refines the Tsetlin Machine's feedback mechanism to enhance accuracy by boosting true positive feedback. This adjustment replaces the standard reward probability $\frac{s - 1}{s}$ with 1.0 and sets the penalty probability $\frac{1}{s}$ to 0.0, significantly strengthening the reinforcement of correct Include actions compared to the standard TM. Additionally, penalties for Exclude actions are adjusted to maintain balance and prevent over-penalization, ensuring the model remains appropriately responsive without becoming overly conservative.

The second modification addresses the treatment of negated features within clauses. Traditional TM structures clauses as conjunctions of both original (non-negated) and negated forms of input features. However, in NTM, the inclusion of negated features ($\bar{I}_{\iota}^{\kappa}$) is disabled. This simplification results in each clause containing only the original forms of input features:
\[ C_{\iota}^{\kappa} = \bigwedge_{k \in I_{\iota}^{\kappa}} x_k. \]

By eliminating negated literals from the clauses, NTM reduces complexity and enhances interpretability, making it easier to interpret the logical rules for model's decisions.\looseness=-1

\subsection{Extract Cluster using Pre-trained BERT based Model}
Let \( D \) represent a document set with unlabeled data. For our case study, we use the AG News dataset, denoted as \( D_{AG} \). This dataset includes news articles that cover a variety of topics. We cluster the unlabeled samples from \( D_{AG} \) using either BERT-based embeddings (K-means clustering) or Top2Vec (Hierarchical Density-based Clustering).  To ensure robustness and generalization, we validate the pre-trained model using AG News across multiple datasets, demonstrating that fixed clusters can effectively work for various test sets.

We use the pre-trained BERT model to obtain embeddings for each document in \( D_{AG} \). Let \( E = \{ e_1, e_2, \ldots, e_n \} \) represent the set of BERT embeddings, where \( e_i \) is the embedding of document \( d_i \). We apply K-means clustering to \( E \), partitioning the embeddings into \( k \) clusters. For our experiments, we set \( k = 200 \), which was found to be optimal. Each document \( d_i \) is assigned a cluster label \( c_i \) from the set \( \{1, 2, \ldots, k\} \).

Alternatively, we use the Top2Vec model by creating jointly embedded document and word vectors using the BERT Sentence Transformer. The BERT Sentence Transformer generates numerous clusters based on the document content. These clusters are then reduced to 200 clusters to ensure fair evaluation, with each document \( d_i \) assigned a cluster label \( c_i \). The next step involves creating sample-cluster pairs to pre-train the NTM. Each document \( d_i \) in \( D_{AG} \) is paired with its cluster label \( c_i \) to form the training set for NTM. Let \( (d_i, c_i) \) denote a sample-cluster pair, where \( d_i \) is the document and \( c_i \) is the cluster label. The set of sample-cluster pairs is represented as \( S = \{ (d_1, c_1), (d_2, c_2), \ldots, (d_n, c_n) \} \).

\subsection{Pre-training Non-Negated Tsetlin Machine}

In this phase, the sample-cluster pairs are trained using the Non-Negated Tsetlin Machine (NTM). Each sample is represented as a Boolean Bag of Words (BOW), with its associated cluster serving as the label. In a typical Tsetlin Machine, each word in the BOW is managed by a pair of Tsetlin Automata (TA): one (\textit{TA}) controls the original form of the literal, and the other (\textit{TA'}) controls its negation. However, in this approach, we disregard the negation, focusing solely on the literals controlled by \textit{TA}, as illustrated in Figure~\ref{fig2}.

During NTM learning, each clause learns a combination of literals in conjunction. Each TA decides whether to include or exclude a literal, operating in two modes: Include or Exclude, with $2N$ states. When a TA moves from state $1$ to $N$, it performs the Exclude action. When it moves from state $N+1$ to $2N$, it performs the Include action. These moves are triggered by feedback in the form of Reward, Penalty, or Inaction; the full Type I and Type II feedback tables are given in Appendix~\ref{app:feedback}.

Once the model is trained, we observe the following: for Cluster 1, positive clauses such as \(C_1\) and \(C_3\) contain words like \textit{basketball}, \textit{olympics}, and \textit{team}, whereas Cluster 2 has clauses that include words like \textit{space}, \textit{observe}, and \textit{camera}. The TA states for these words, after reaching optimal training epochs, indicate their confidence levels within each cluster. For instance, words like \textit{space} and \textit{olympics} are in deeper TA states, making them more repetitive across clauses compared to other words.

This confidence state gives the Tsetlin Machine an advantage over models like Top2Vec, whose cluster keywords are more generic. We define the confidence of literal $j$ for class $y$ as $\text{Conf}_y(j) = \sum_{i \in C_y^{(p)}} \text{TA}_{ij}$, where $C_y^{(p)}$ are the positive (odd-indexed) clauses of class $y$ that include literal $j$ and $\text{TA}_{ij}$ is the state of the TA including $j$ in clause $i$; a deeper state implies higher confidence.

For example, Cluster 1 includes words like \textit{olympic}, \textit{basketball}, and \textit{teams}, indicating a sports-related cluster, whereas Cluster 2 includes words like \textit{space}, \textit{camera}, and \textit{observe}, indicating a space or sci-fi cluster. This method generates an interpretable confidence-based word representation for clusters obtained from BERT or Top2Vec models. This representation is then used as pre-trained explainable information for fine-tuning the Tsetlin Machine or any other interpretable model.

\subsection{Fine-Tune with Pre-trained Word Representation}
In the fine-tuning stage, a labeled dataset is utilized. Similar to the pretraining stage, BERT-variants or Top2Vec are applied to assign clusters to the text samples. For each cluster, the Bag of Words (BOW) representation of the text samples is augmented with the pre-trained keywords obtained from the NTM. 
The enriched BOW representations are then used to fine-tune the vanilla TM. This integration of pre-trained keywords aims to leverage the interpretable and discriminative power of the NTM, thereby improving the performance and interpretability of the vanilla TM in handling the labeled dataset. By combining pre-trained word representations with fine-tuning, we aim to create a more robust and interpretable TM model, capable of accurately capturing and leveraging the nuances of the text data within each cluster.

\section{Experiments and Results}
In this section, we start by introducing the benchmark datasets used to evaluate our model. Our goal was to include widely recognized topic classification datasets. Following this, we introduce several baseline models that serve as benchmarks for comparison with our approach. Finally, we analyze our proposed model to demonstrate its capabilities and performance consistency across different scenarios.

\noindent\textbf{Datasets.} We evaluate on five topic classification datasets: NYT-Topics, derived from the New York Times Annotated Corpus \cite{sandhaus2008nytac,Meng2019DiscriminativeTM}, AG-News \cite{Zhang2015CharacterlevelCN}, DBpedia \cite{Lehmann2015DBpediaA}, and the R8 and R52 subsets of Reuters-21578 (8 and 52 categories, respectively). Dataset statistics and curation details are provided in Appendix~\ref{app:datasets}.

\noindent\textbf{Baselines.} We compare against a curated set of baselines spanning traditional and neural methods: BoW with TF-IDF \cite{SprckJones2021ASI}, char-CNN \cite{Zhang2015CharacterlevelCN}, LSTM with and without GloVe \cite{Hochreiter1997LongSM}, FastText \cite{joulin-etal-2017-bag}, BERT-base and BERT-large \cite{Sun2019HowTF,chen-miyake-2021-label}, vanilla TM and GloVe-embedded TM \cite{yadav-etal-2021-enhancing}, and Top2Vec \cite{Angelov2020Top2VecDR} for keyword comparison. Rather than enumerating many variants of each model, we keep one representative per family so that the comparison isolates the effect of interpretable semantic pre-training rather than architectural tuning. Detailed descriptions of each baseline are provided in Appendix~\ref{app:baselines}.

\begin{table*}[t]
  \caption{Performance of fine-tuned Vanilla TM on pre-trained cluster representation from NTM. The baseline ranges from typical BOW to BERT-based models. }
  \label{tab:sota}
  \centering
  \scalebox{0.85}{
  \begin{tabular}{l|c|c|c|c|c}
    \toprule
    \textbf{Model/Datasets} & \textbf{AG-News} & \textbf{NYT} & \textbf{DBpedia} & \textbf{R8} & \textbf{R52}\\
    \midrule
    BoW TFIDF & 89.64 & 88.24 & 97.37 & 93.74 &  86.95 \\
    char-CNN & 87.20 & 88.67 & 98.3 & 94.02 & 85.37 \\
    LSTM & 86.06 & 87.72 & 98.55 &  93.68 &  85.54 \\
    LSTM (GloVe) & 92.10 & 89.54 & 98.70 &  96.09 & 90.48 \\
    fastText, h = 10  & 91.50 & 91.76 &  98.10 &  94.74 &  90.99 \\
    BERT-base & 94.75 & 93.83 & 99.29 & 97.49 & 94.26 \\
    BERT-large & 95.71 & 94.82 & 99.38 & 98.28 & 94.32 \\
    TM & 88.34 & 88.50 & 98.20 &  96.16 & 84.62 \\
    TM (GloVe) & 90.12 & 90.32 & 98.42 & 97.50 &  89.14 \\
    \midrule
    \textbf{TM (NTM pretrain with Top2vec)} & 92.40 ± 0.92 & 92.32 ± 0.92 & 98.66 ± 0.92 & 97.70 ± 0.92 & 93.24 ± 0.92 \\
    \textbf{TM (NTM pretrain with BERT-base)} & 94.12 ± 0.71 & 93.41 ± 0.71 & 98.98 ± 0.71 & 97.88 ± 0.71 & 95.44 ± 0.71 \\
    \textbf{TM (NTM pretrain with BERT-large)} & \underline{\textbf{94.80}} ± 0.64 & \underline{\textbf{93.72}} ± 0.64 & \underline{\textbf{99.10}} ± 0.64 & \underline{\textbf{97.92}} ± 0.64 & \textbf{95.27} ± 0.64 \\
    \bottomrule
  \end{tabular}}
\end{table*}

\subsection{Experimental Details}

For pre-training we obtain embeddings from BERT (base and large) and from Top2Vec initialized with the Universal Sentence Encoder, and cluster the unlabeled AG-News samples into $\{50, 100, 200, 300, 400, 500\}$ clusters. The NTM uses $2000$ clauses ($1000$ per cluster), threshold $T=4000$, and specificity $s=10$, with the weighted Tsetlin Machine of \citet{Abeyrathna2020ExtendingTT}. During fine-tuning we reintroduce negated literals and lower $s$ to $5$ for more flexible literal inclusion. Full hyperparameter settings and their rationale are provided in Appendix~\ref{app:details}.

\subsection{Performance Comparison}
Table \ref{tab:sota} presents a comprehensive comparison of the accuracy percentages across several datasets, highlighting the performance of various models, including traditional methods from BOW , LSTM, CNN, BERT-based models, and our proposed Tsetlin Machine (TM) approach enhanced with pre-trained semantic clusters. While BERT-based models, particularly BERT-large, tend to achieve the highest accuracy, our approach, pre-training TM with BERT-derived clusters, consistently delivers competitive performance, often within 1-2\% of BERT-large, and in some cases, even surpasses it.

Across the datasets, TM pre-trained with BERT clusters typically achieves accuracy within 1-2\% of BERT-large. For example, on the AG-News and NYT datasets, our method trails BERT-large by only about 0.9\% and 1.1\%, respectively, while on the DBpedia dataset, the difference is a mere 0.3\%. This demonstrates that the semantic richness captured by BERT can be effectively transferred into TM, significantly boosting its accuracy. Moreover, our method outperforms TM initialized with Top2Vec by 1-2\% on most datasets, further validating the advantage of using keyword extracted from NTM in contrast with Top2Vec. 

Although there is a slight performance gap compared to BERT-large, this is largely due to the information loss inherent in the retraining process within TM, where the binary nature of TM's input representation leads to some loss of the nuanced semantic information captured by BERT. Despite this, our method achieves impressive accuracy, within 1\% of BERT-large on the DBpedia and R52 datasets, demonstrating that this trade-off is minimal when considering the increased interpretability and explainability offered by TM.

Overall, the results strongly support our main hypothesis: pre-training TM with semantically enriched clusters from a pre-trained language model like BERT significantly enhances its performance, making it a highly competitive and interpretable alternative to more complex and less transparent models. This approach successfully narrows the accuracy gap between TM and state-of-the-art BERT models, setting a new standard for explainable text classification.

\subsection{Impact of Cluster Sizes on Performance}
We first study the impact of selecting various cluster sizes for BERT and Top2Vec approaches. We run the selected cluster sizes with the same configuration selected above for NTM and fine-tuning vanilla TM. For ease of illustration, we use only three datasets to show the cluster size impact. The performance of these cluster sizes is shown in Table \ref{tab:cluster_size}.

From the results, we observe that smaller cluster sizes (e.g., $50$) provide limited contextual information, resulting in lower accuracy across all models. For instance, in AG-News, Top2Vec achieves 90.24\%, BERT-b 91.50\%, and BERT-l 92.52\%. As the cluster size increases to $100$, the accuracy improves due to enhanced contextual information. For example, in AG-News, Top2Vec achieves 91.72\%, BERT-b 93.20\%, and BERT-l 93.87\%. The cluster size of 200 provides the highest accuracy across all datasets and models, with AG-News showing Top2Vec at 92.40\%, BERT-b at 94.12\%, and BERT-l at 94.80\%. This indicates that 200 is an optimal cluster size for this setup. However, increasing the cluster size beyond $200$ results in a decline in accuracy. This is due to the disintegration of contextual information into smaller chunks, making it harder for the NTM to learn effectively. For instance, in AG-News, Top2Vec drops to 91.22\% at a cluster size of 300, 90.13\% at 400, and 88.88\% at $500$. In summary, a cluster size of $200$ strikes the best balance between contextual richness and coherence in all the selected datasets, while sizes smaller or larger than $200$ tend to reduce model performance.


\begin{table*}[t]
  \caption{Accuracy of Different Models on Various Datasets for Different Cluster Sizes}
  \label{tab:cluster_size}
  \centering
  \scalebox{0.88}{
  \begin{tabular}{l|c|c|c|c|c|c|c|c|c|c|c}
    \toprule
    \textbf{Cluster Size} & \multicolumn{3}{c|}{\textbf{AG-News}} & \multicolumn{3}{c|}{\textbf{DBpedia}} & \multicolumn{3}{c|}{\textbf{R8}} \\
    \cmidrule{2-10}
    & \textbf{Top2Vec} & \textbf{BERT-b} & \textbf{BERT-l} & \textbf{Top2Vec} & \textbf{BERT-b} & \textbf{BERT-l} & \textbf{Top2Vec} & \textbf{BERT-b} & \textbf{BERT-l} \\
    \midrule
    \textbf{50} & 90.24 & 91.50 & 92.52 & 97.50 & 97.58 & 97.89 & 95.89 & 96.28 & 96.70 \\
    \textbf{100} & \underline{\textbf{91.72}} & 93.20 & \underline{\textbf{93.87}} & \underline{\textbf{98.28}} & \underline{\textbf{98.48}} & 98.80 & \underline{\textbf{96.40}} & \underline{\textbf{97.14}} & \underline{\textbf{97.42}} \\
    \textbf{200} & \textbf{92.40} & \textbf{94.12} & \textbf{94.80} & \textbf{98.66} & \textbf{98.98} & \textbf{99.10} & \textbf{97.70} & \textbf{97.88} & \textbf{97.92} \\
    \textbf{300} & 91.22 & \underline{\textbf{93.50}} & 93.01 & 97.82 & 97.96 & \underline{\textbf{98.92}} & 96.35 & 96.96 & 97.20 \\
    \textbf{400} & 90.13 & 92.00 & 92.75 & 97.42 & 97.42 & 97.88 & 95.90 & 96.10 & 96.35 \\
    \textbf{500} & 88.88 & 89.70 & 90.82 & 96.12 & 96.30 & 96.68 & 94.26 & 94.80 & 95.10 \\
    \bottomrule
  \end{tabular}}
\end{table*}
\subsection{Impact Confidence of keywords on Performance}
\begin{figure}[t]
\centering
\includegraphics[width=0.8\columnwidth]{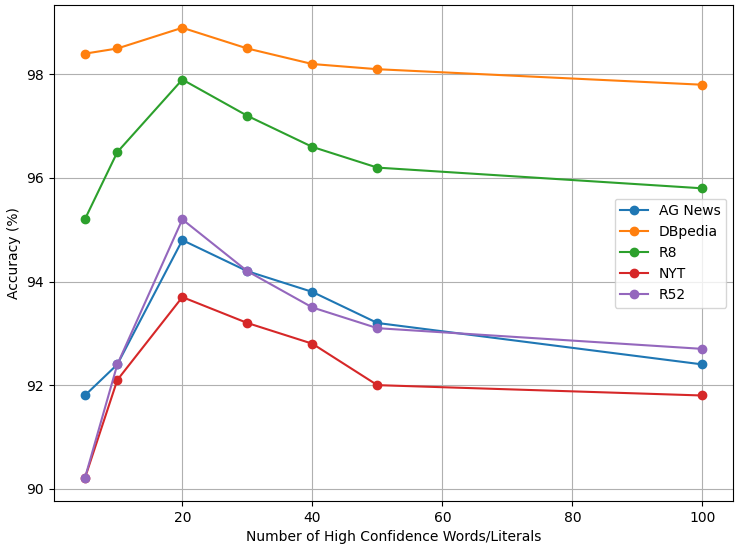} 
\caption{Comparison of accuracy for different selected number of high confidence words/literals based on the sum of their TA states in the clause. We have used BERT-l and $200$ cluster size.}
\label{fig3}
\end{figure}

Additionally, we demonstrate the effectiveness of confidence-based keywords derived from the NTM TA-state confidence measure defined above. By selecting high-confidence keywords based on the states of Tsetlin Automata (TA) associated with each word, we enhance the impact of pre-trained information during fine-tuning. This highlights the advantage of using NTM pre-trained information over Top2Vec topic keywords. We experiment by using several number of high confidence literals for fine-tuning it on vanilla TM as shown in Figure~\ref{fig3}.\looseness=-1

As illustrated in Figure~\ref{fig3}, the performance metrics for AG News, DBpedia, R8, NYT, and R52 datasets exhibit a general improvement with the increase in the number of high-confidence words/literals. For instance, AG News shows a peak performance at 20 high-confidence words/literals, reaching 94.8\%, and then slightly declines. Similarly, DBpedia achieves its highest performance of 98.9\% at 20 high-confidence words/literals. R8 also peaks at 20 words with 97.9\% performance. These trends suggest that there is an optimal number of high-confidence words/literals that maximizes performance for each dataset.

Beyond this point, the benefit diminishes, indicating the importance of selecting an appropriate number of high-confidence words/literals for fine-tuning the model. This is because, even in NTM pre-trained information, there exist some noisy words that are out of context for the cluster type. If we add more words, the chances that the context will include noisy or out-of-context words increase, thereby reducing accuracy.

\section{Explainability in NTM/Fine-tuned TM}
\label{sec:explain}
\begin{figure}[t]
\centering
\includegraphics[width=0.85\columnwidth]{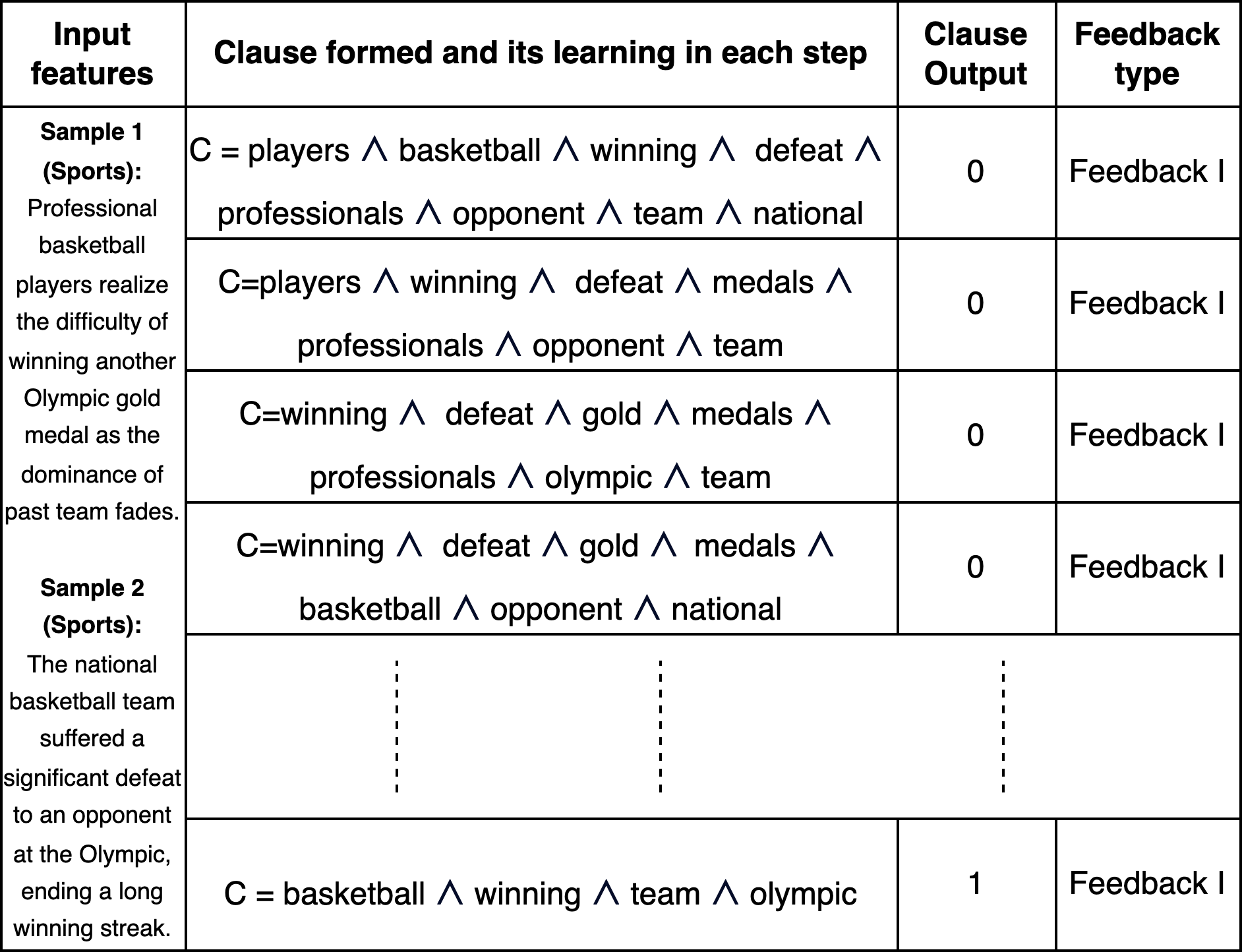}
\caption{Learning Process of pre-training NTM without negated feature and generating cluster representation with high confidence TA states.}
\label{fig4}
\end{figure}

We illustrate the transparency of both phases on a sports-cluster example (Figure~\ref{fig4}). During pre-training, the NTM clauses start from random literal patterns and, guided by Type I feedback, converge to a monotone pattern for the sports cluster, [basketball AND winning AND team AND olympic], without any negated literals. During fine-tuning with the vanilla TM, negated literals are reintroduced. A labeled sports sentence whose obvious cues ([game, soccer, team, victory]) give only a thin representation would otherwise force the clause to rely heavily on negated features; embedding the cluster descriptors enriches the input with semantic words such as [basketball, winning, olympic], yielding a representation that depends on only a few negations and improves the model's understanding of the class, similar to BERT-style embeddings. A full step-by-step trace of clause convergence and feature enrichment is provided in Appendix~\ref{app:explain}.

\section{Conclusion}
In this paper, we introduced an innovative method to enhance the Tsetlin Machine (TM) by pre-training it with semantic clusters derived from language models like BERT. By clustering text data using K-means or Top2Vec and leveraging these clusters to pre-train the TM, we effectively transferred rich semantic information into the TM. This approach significantly improved the TM's performance in text classification tasks while maintaining its interpretability. Our results demonstrate that the combination of semantic clustering and TM pre-training provides a powerful and explainable model, offering a promising direction for our future research and applications in explainable AI.

\clearpage
\section*{Limitations}

The proposed method depends on the quality of semantic clusters generated by a pre-trained language model, which may affect performance on noisy or low-resource data. The pre-training stage adds offline computational cost, and fine-grained contextual dependencies may remain better captured by fully neural models.



\bibliography{custom}

@inproceedings{Devlin2019BERTPO,
  title={BERT: Pre-training of Deep Bidirectional Transformers for Language Understanding},
  author={Jacob Devlin and Ming-Wei Chang and Kenton Lee and Kristina Toutanova},
  booktitle={North American Chapter of the Association for Computational Linguistics},
  year={2019},
}

@article{Raffel2019ExploringTL,
  title={Exploring the Limits of Transfer Learning with a Unified Text-to-Text Transformer},
  author={Colin Raffel and Noam M. Shazeer and Adam Roberts and Katherine Lee and Sharan Narang and Michael Matena and Yanqi Zhou and Wei Li and Peter J. Liu},
  journal={J. Mach. Learn. Res.},
  year={2019},
  volume={21},
  pages={140:1-140:67},

}

@inproceedings{10.5555/3495724.3495883,
author = {Brown, Tom B. and Mann, Benjamin and Ryder, Nick and Subbiah, Melanie and Kaplan, Jared and Dhariwal, Prafulla and Neelakantan, Arvind and Shyam, Pranav and Sastry, Girish and Askell, Amanda and Agarwal, Sandhini and Herbert-Voss, Ariel and Krueger, Gretchen and Henighan, Tom and Child, Rewon and Ramesh, Aditya and Ziegler, Daniel M. and Wu, Jeffrey and Winter, Clemens and Hesse, Christopher and Chen, Mark and Sigler, Eric and Litwin, Mateusz and Gray, Scott and Chess, Benjamin and Clark, Jack and Berner, Christopher and McCandlish, Sam and Radford, Alec and Sutskever, Ilya and Amodei, Dario},
title = {Language models are few-shot learners},
year = {2020},
isbn = {9781713829546},
publisher = {Curran Associates Inc.},
address = {Red Hook, NY, USA},
booktitle = {Proceedings of the 34th International Conference on Neural Information Processing Systems},
articleno = {159},
numpages = {25},
series = {NIPS '20}
}

@inproceedings{10.5555/3524938.3525306,
author = {Guu, Kelvin and Lee, Kenton and Tung, Zora and Pasupat, Panupong and Chang, Ming-Wei},
title = {REALM: retrieval-augmented language model pre-training},
year = {2020},
publisher = {JMLR.org},
booktitle = {Proceedings of the 37th International Conference on Machine Learning},
articleno = {368},
numpages = {10},
series = {ICML'20}
}

@article{Liu2019FinetuneBF,
  title={Fine-tune BERT for Extractive Summarization},
  author={Yang Liu},
  journal={ArXiv},
  year={2019},
  volume={abs/1903.10318},

}

@inproceedings{wadden-etal-2019-entity,
    title = "Entity, Relation, and Event Extraction with Contextualized Span Representations",
    author = "Wadden, David  and
      Wennberg, Ulme  and
      Luan, Yi  and
      Hajishirzi, Hannaneh",
    editor = "Inui, Kentaro  and
      Jiang, Jing  and
      Ng, Vincent  and
      Wan, Xiaojun",
    booktitle = "Proceedings of the 2019 Conference on Empirical Methods in Natural Language Processing and the 9th International Joint Conference on Natural Language Processing (EMNLP-IJCNLP)",
    month = nov,
    year = "2019",
    address = "Hong Kong, China",
    publisher = "Association for Computational Linguistics",
    url = "https://aclanthology.org/D19-1585",
    doi = "10.18653/v1/D19-1585",
    pages = "5784--5789",

}

@inproceedings{ZhuXWHQZLL20,
  author = {Zhu, Jinhua and Xia, Yingce and Wu, Lijun and He, Di and Qin, Tao and Zhou, Wengang and Li, Houqiang and Liu, Tie-Yan},
  booktitle = {ICLR},
  title = {Incorporating BERT into Neural Machine Translation.},
  year = 2020
}

@inproceedings{Wang2018GLUEAM,
  title={GLUE: A Multi-Task Benchmark and Analysis Platform for Natural Language Understanding},
  author={Alex Wang and Amanpreet Singh and Julian Michael and Felix Hill and Omer Levy and Samuel R. Bowman},
  booktitle={BlackboxNLP@EMNLP},
  year={2018},

}

@inproceedings{10.5555/3454287.3454581,
author = {Wang, Alex and Pruksachatkun, Yada and Nangia, Nikita and Singh, Amanpreet and Michael, Julian and Hill, Felix and Levy, Omer and Bowman, Samuel R.},
title = {SuperGLUE: a stickier benchmark for general-purpose language understanding systems},
year = {2019},
publisher = {Curran Associates Inc.},
address = {Red Hook, NY, USA},
booktitle = {Proceedings of the 33rd International Conference on Neural Information Processing Systems},
articleno = {294},
numpages = {15}
}

@inproceedings{jain-wallace-2019-attention,
    title = "{A}ttention is not {E}xplanation",
    author = "Jain, Sarthak  and
      Wallace, Byron C.",
    editor = "Burstein, Jill  and
      Doran, Christy  and
      Solorio, Thamar",
    booktitle = "Proceedings of the 2019 Conference of the North {A}merican Chapter of the Association for Computational Linguistics: Human Language Technologies, Volume 1 (Long and Short Papers)",
    month = jun,
    year = "2019",
    address = "Minneapolis, Minnesota",
    publisher = "Association for Computational Linguistics",
    doi = "10.18653/v1/N19-1357",
    pages = "3543--3556",

}

@article{Rudin2018StopEB,
  title={Stop explaining black box machine learning models for high stakes decisions and use interpretable models instead},
  author={Cynthia Rudin},
  journal={Nature Machine Intelligence},
  year={2018},
  volume={1},
  pages={206 - 215},

}

@article{wu-etal-2023-transparency,
    title = "Transparency Helps Reveal When Language Models Learn Meaning",
    author = "Wu, Zhaofeng  and
      Merrill, William  and
      Peng, Hao  and
      Beltagy, Iz  and
      Smith, Noah A.",
    journal = "Transactions of the Association for Computational Linguistics",
    volume = "11",
    year = "2023",
    address = "Cambridge, MA",
    pages = "617--634",

}

@inbook{molnar2022interpretable,
  title={Interpretable Machine Learning},
  author={Christoph Molnar},
  year={2022},
  url={https://christophmolnar.com/books/interpretable-machine-learning/},

}

@inproceedings{10.24963/ijcai.2023/378,
author = {Abeyrathna, K. Darshana and Abouzeid, Ahmed A. O. and Bhattarai, Bimal and Giri, Charul and Glimsdal, Sondre and Granmo, Ole-Christoffer and Jiao, Lei and Saha, Rupsa and Sharma, Jivitesh and Tunheim, Svein A. and Zhang, Xuan},
title = {Building concise logical patterns by constraining tsetlin machine clause size},
year = {2023},
booktitle = {Proceedings of the Thirty-Second International Joint Conference on Artificial Intelligence},
articleno = {378},
numpages = {9},
location = {<conf-loc>, <city>Macao</city>, <country>P.R.China</country>, </conf-loc>},
series = {IJCAI '23}
}

@inproceedings{yadav-etal-2021-enhancing,
    title = "Enhancing Interpretable Clauses Semantically using Pretrained Word Representation",
    author = "Yadav, Rohan Kumar  and
      Jiao, Lei  and
      Granmo, Ole-Christoffer  and
      Goodwin, Morten",
    editor = "Bastings, Jasmijn  and
      Belinkov, Yonatan  and
      Dupoux, Emmanuel  and
      Giulianelli, Mario  and
      Hupkes, Dieuwke  and
      Pinter, Yuval  and
      Sajjad, Hassan",
    booktitle = "Proceedings of the Fourth BlackboxNLP Workshop on Analyzing and Interpreting Neural Networks for NLP",
    year = "2021",
    address = "Punta Cana, Dominican Republic",
    publisher = "Association for Computational Linguistics",
    pages = "265--274",

}

@article{Granmo2018TheTM,
  title={The Tsetlin Machine - A Game Theoretic Bandit Driven Approach to Optimal Pattern Recognition with Propositional Logic},
  author={Ole-Christoffer Granmo},
  journal={ArXiv},
  year={2018},
  volume={abs/1804.01508},
  url={https://api.semanticscholar.org/CorpusID:4597378}
}

@inproceedings{bhattarai-etal-2022-explainable,
    title = "Explainable Tsetlin Machine Framework for Fake News Detection with Credibility Score Assessment",
    author = "Bhattarai, Bimal  and
      Granmo, Ole-Christoffer  and
      Jiao, Lei",
    editor = "Calzolari, Nicoletta  and
      B{\'e}chet, Fr{\'e}d{\'e}ric  and
      Blache, Philippe  and
      Choukri, Khalid  and
      Cieri, Christopher  and
      Declerck, Thierry  and
      Goggi, Sara  and
      Isahara, Hitoshi  and
      Maegaard, Bente  and
      Mariani, Joseph  and
      Mazo, H{\'e}l{\`e}ne  and
      Odijk, Jan  and
      Piperidis, Stelios",
    booktitle = "Proceedings of the Thirteenth Language Resources and Evaluation Conference",
    year = "2022",
    address = "Marseille, France",
    publisher = "European Language Resources Association",
    pages = "4894--4903",
}

@inproceedings{Yadav2021HumanLevelIL,
  title={Human-Level Interpretable Learning for Aspect-Based Sentiment Analysis},
  author={Rohan Kumar Yadav and Lei Jiao and Ole-Christoffer Granmo and Morten Goodwin Olsen},
  booktitle={AAAI Conference on Artificial Intelligence},
  year={2021},
  url={https://api.semanticscholar.org/CorpusID:235363658}
}

@article{Berge2018UsingTT,
  title={Using the Tsetlin Machine to Learn Human-Interpretable Rules for High-Accuracy Text Categorization With Medical Applications},
  author={Geir Thore Berge and Ole-Christoffer Granmo and Tor Oddbj{\o}rn Tveit and Morten Goodwin and Lei Jiao and Bernt Viggo Matheussen},
  journal={IEEE Access},
  year={2018},
  volume={7},
  pages={115134-115146},
  url={https://api.semanticscholar.org/CorpusID:52195410}
}

@article{Saha2022InterpretableTC,
  title={Interpretable Text Classification in Legal Contract Documents using Tsetlin Machines},
  author={Rupsa Saha and Sander Jyhne},
  journal={2022 International Symposium on the Tsetlin Machine (ISTM)},
  year={2022},
  pages={7-12},
  url={https://api.semanticscholar.org/CorpusID:253123132}
}

@article{Angelov2020Top2VecDR,
  title={Top2Vec: Distributed Representations of Topics},
  author={Dimitar Angelov},
  journal={ArXiv},
  year={2020},
  volume={abs/2008.09470},
  url={https://api.semanticscholar.org/CorpusID:221246303}
}

@misc{sandhaus2008nytac,
  author = {Evan Sandhaus},
  title = {The New York Times Annotated Corpus},
  year = {2008},
  organization = {Linguistic Data Consortium},
  publisher = {University of Pennsylvania},
}

@article{Meng2019DiscriminativeTM,
  title={Discriminative Topic Mining via Category-Name Guided Text Embedding},
  author={Yu Meng and Jiaxin Huang and Guangyuan Wang and Zihan Wang and Chao Zhang and Yu Zhang and Jiawei Han},
  journal={Proceedings of The Web Conference 2020},
  year={2019},
  url={https://api.semanticscholar.org/CorpusID:210932696}
}

@inproceedings{Zhang2015CharacterlevelCN,
  title={Character-level Convolutional Networks for Text Classification},
  author={Xiang Zhang and Junbo Jake Zhao and Yann LeCun},
  booktitle={Neural Information Processing Systems},
  year={2015},
  url={https://api.semanticscholar.org/CorpusID:368182}
}

@article{Lehmann2015DBpediaA,
  title={DBpedia - A large-scale, multilingual knowledge base extracted from Wikipedia},
  author={Jens Lehmann and Robert Isele and Max Jakob and Anja Jentzsch and Dimitris Kontokostas and Pablo N. Mendes and Sebastian Hellmann and Mohamed Morsey and Patrick van Kleef and S. Auer and Christian Bizer},
  journal={Semantic Web},
  year={2015},
  volume={6},
  pages={167-195},
  url={https://api.semanticscholar.org/CorpusID:1181640}
}

@article{SprckJones2021ASI,
  title={A statistical interpretation of term specificity and its application in retrieval},
  author={Karen Sp{\"a}rck Jones},
  journal={J. Documentation},
  year={2021},
  volume={60},
  pages={493-502},
  url={https://api.semanticscholar.org/CorpusID:2996187}
}

@article{Hochreiter1997LongSM,
  title={Long Short-Term Memory},
  author={Sepp Hochreiter and J{\"u}rgen Schmidhuber},
  journal={Neural Computation},
  year={1997},
  volume={9},
  pages={1735-1780},
  url={https://api.semanticscholar.org/CorpusID:1915014}
}

@inproceedings{joulin-etal-2017-bag,
    title = "Bag of Tricks for Efficient Text Classification",
    author = "Joulin, Armand  and
      Grave, Edouard  and
      Bojanowski, Piotr  and
      Mikolov, Tomas",
    booktitle = "Proceedings of the 15th Conference of the {E}uropean Chapter of the Association for Computational Linguistics: Volume 2, Short Papers",

    year = "2017",
    address = "Valencia, Spain",
    publisher = "Association for Computational Linguistics",
    pages = "427--431"
}

@inproceedings{Sun2019HowTF,
  title={How to Fine-Tune BERT for Text Classification?},
  author={Chi Sun and Xipeng Qiu and Yige Xu and Xuanjing Huang},
  booktitle={China National Conference on Chinese Computational Linguistics},
  year={2019},
  url={https://api.semanticscholar.org/CorpusID:153312532}
}

@inproceedings{chen-miyake-2021-label,
    title = "Label-Guided Learning for Item Categorization in e-Commerce",
    author = "Chen, Lei  and
      Miyake, Hirokazu",
    booktitle = "Proceedings of the 2021 Conference of the North American Chapter of the Association for Computational Linguistics: Human Language Technologies: Industry Papers",

    year = "2021",
    address = "Online",
    publisher = "Association for Computational Linguistics",
    pages = "296--303"
}

@article{Abeyrathna2020ExtendingTT,
  title={Extending the Tsetlin Machine With Integer-Weighted Clauses for Increased Interpretability},
  author={Kuruge Darshana Abeyrathna and Ole-Christoffer Granmo and Morten Goodwin},
  journal={IEEE Access},
  year={2020},
  volume={9},
  pages={8233-8248},
  url={https://api.semanticscholar.org/CorpusID:218581474}
}

\appendix

\section{Extended Related Work}
\label{app:related}

Pre-trained language models are most commonly used by fine-tuning a pre-trained model on a supervised downstream dataset: most of the model's parameters are retained while a randomly initialized task-specific layer is added to adapt the model to the new task \cite{10.5555/3524938.3525306,Liu2019FinetuneBF,wadden-etal-2019-entity,ZhuXWHQZLL20}. This paradigm has substantially improved natural language understanding across many tasks and industrial applications. Despite this success, transparency and explainability remain open problems, particularly in sensitive fields where the decision process must be understood. Transparent models provide more insight into their inner workings but frequently trade away accuracy, which motivates models such as the Tsetlin Machine that aim to retain both. Earlier work integrated GloVe embeddings alongside BOW features in TM and reached accuracy comparable to GloVe-initialized LSTM and CNN models \cite{yadav-etal-2021-enhancing}; our work instead transfers contextual knowledge from transformer-based models such as BERT and from Top2Vec \cite{Angelov2020Top2VecDR} through semantic clusters, while keeping the representation interpretable.

\section{Dataset Details}
\label{app:datasets}

We focus on topic classification to assess general text classification performance. NYT-Topics is derived from the New York Times Annotated Corpus spanning January 1987 to June 2007, which contains 1.8 million articles with metadata \cite{sandhaus2008nytac}. The NYT-Topics subset was curated by selecting major categories (each containing over 100 documents) from topics and locations, keeping documents that carry a single label for both categories so that each document has exactly one ground-truth topic label and one ground-truth location label \cite{Meng2019DiscriminativeTM}. We additionally use AG-News \cite{Zhang2015CharacterlevelCN}, DBpedia \cite{Lehmann2015DBpediaA}, and the R8 and R52 subsets of Reuters-21578, segmented into 8 and 52 categories respectively. Statistics for all datasets are summarized in Table~\ref{tab:datasets}.

\begin{table}[h]
  \caption{Statistics of selected datasets.}
  \label{tab:datasets}
  \centering
  \scalebox{0.75}{%
    \begin{tabular}{l|c|c|c}
      \toprule
      \textbf{Datasets} & \textbf{Train} & \textbf{Test} & \textbf{Class} \\
      \midrule
      AG-News & 30,000 & 7,600 & 4  \\
      NYT-Topic & 30,000 & 3,000 & 9  \\
      DBPedia & 30,000 & 70,000 & 14 \\
      R8 &  5,485 &  2,189 & 8 \\
      R52 & 6,532  & 2,568 & 52 \\
      \bottomrule
    \end{tabular}%
  }
\end{table}

\section{Tsetlin Machine Feedback}
\label{app:feedback}

Table~\ref{table_combined} details the Type I and Type II feedback that drive the Tsetlin Automata during NTM learning. Type I feedback reinforces true positive outputs and combats false negatives, while Type II feedback combats false positives. The probabilities are governed by the specificity hyperparameter $s$.

\begin{table}[t]
  \caption{The Type I and Type II Feedback.}
  \label{table_combined}
  \centering
  \scalebox{0.55}{
  \begin{tabular}{l|l|l|l|l|l|l}
    \toprule
    \multirow{2}{*}{Input} & \multicolumn{3}{c|}{Type I Feedback} & \multicolumn{3}{c}{Type II Feedback} \\
    \cmidrule{2-7}
    & Clause & \ \ \ \ \ \ \ 1 & \ \ \ \ \ \ \ 0 & Clause & \ \ \ \ \ \ \ 1 & \ \ \ \ \ \ \ 0 \\
    \cmidrule{2-7}
    & Literal & \ \ 1 \ \ \ \ \ \ 0 & \ \ 1 \ \ \ \ \ \ 0 & Literal & \ \ 1 \ \ \ \ \ \ 0 & \ \ 1 \ \ \ \ \ \ 0 \\
    \midrule
    \multirow{2}{*}{Include Literal} & P(Reward) & $\frac{s-1}{s}$\ \ \ NA & \ \ 0 \ \ \ \ \ \ 0 & P(Reward) & \ \ 0 \ \ \ NA & \ \ 0 \ \ \ \ \ \ 0 \\ [1mm]
    & P(Inaction) & $\ \ \frac{1}{s}$\ \ \ \ \ NA & $\frac{s-1}{s}$ \ $\frac{s-1}{s}$ & P(Inaction) & 1.0 \ \  NA &  1.0 \ \ \ 1.0 \\ [1mm]
    & P(Penalty) & \ \ 0 \ \ \ \ \ NA & $\ \ \frac{1}{s}$ \ \ \ \ \  $\frac{1}{s}$ & P(Penalty) & \ \ 0 \ \ \ NA & \ \ 0 \ \ \ \ \ \ 0 \\ [1mm]
    \midrule
    \multirow{2}{*}{Exclude Literal} & P(Reward) & \ \ 0 \ \ \ \ \ \ $\frac{1}{s}$ & $\ \ \frac{1}{s}$ \ \ \ \ \ $\frac{1}{s}$ & P(Reward) & \ \ 0 \ \ \ \ 0 & \ \ 0 \ \ \ \ \ \ 0 \\ [1mm]
    & P(Inaction) & $ \ \ \frac{1}{s}$\ \ \ \ $\frac{s-1}{s}$  & $\frac{s-1}{s}$ \ $\frac{s-1}{s}$ & P(Inaction) & 1.0 \ \ \ 0 &  1.0 \ \ \ 1.0 \\ [1mm]
    & P(Penalty) & $\frac{s-1}{s}$ \ \ \ \ 0 & \ \ 0 \ \ \ \ \ \ 0 & P(Penalty) & \ \ 0 \ \  1.0 & \ \ 0 \ \ \ \ \ \ 0 \\ [1mm]
    \bottomrule
  \end{tabular}}
\end{table}

\section{Experimental Details}
\label{app:details}

For the Top2Vec approach, which generates as many clusters as possible based on the data type, we reduce the clusters to the selected numbers using the Top2Vec library. The NTM uses $2000$ clauses, that is, $1000$ clauses per cluster, to capture rich AND rules for learning semantics. The threshold $T=4000$ means clauses stop updating once the clause sum for each cluster reaches $T$. The specificity $s=10$ controls the inclusion of literals in the clauses; this value gives a reasonable probability of selecting literals, facilitating the transition between states toward a local minimum. We use the weighted Tsetlin Machine throughout \cite{Abeyrathna2020ExtendingTT}. For fine-tuning, we keep the same configuration but reintroduce negated literals to capture a broader range of patterns and reduce $s$ to $5$ for more flexible literal inclusion, allowing the model to leverage pre-trained knowledge while adapting to the new data.

\section{Baseline Descriptions}
\label{app:baselines}

We provide detailed descriptions of the baselines used in our experiments.

\begin{itemize}
    \item \textbf{BoW TFIDF} selects the most frequent words from each dataset's training subset, using word counts as features. In the TF-IDF version \cite{SprckJones2021ASI}, term frequencies are computed and features are normalized based on inverse document frequency, scaling them by the largest feature value. For TM, this reduces to a Boolean BOW.
    \item \textbf{char-CNN} uses character-level convolutional networks (ConvNets) for text classification. Large-scale datasets were constructed to show that character-level ConvNets can achieve state-of-the-art or competitive results \cite{Zhang2015CharacterlevelCN}.
    \item \textbf{LSTM and LSTM (GloVe)} represent the entire text using the last hidden state \cite{Hochreiter1997LongSM}. We test this model both with and without pre-trained word embeddings.
    \item \textbf{FastText} averages word features to create effective sentence representations, achieving competitive performance with deep learning methods while being much faster \cite{joulin-etal-2017-bag}.
    \item \textbf{BERT}: We fine-tune BERT-base and BERT-large on the selected datasets. For AG-News and DBpedia the baselines are taken from \cite{Sun2019HowTF}. For R8 and R52 we add the LguidedLearn BERT variant, a label-guided learning framework for text classification applied to BERT \cite{chen-miyake-2021-label}.
    \item \textbf{TM}: We compare against the vanilla TM and the GloVe-embedded TM, a typical baseline for such comparisons \cite{yadav-etal-2021-enhancing}.
    \item \textbf{Top2Vec} \cite{Angelov2020Top2VecDR} is used to compare the keywords extracted from NTM with those from Top2Vec; we extract a similar number of keywords per cluster to facilitate this comparison.
\end{itemize}

\section{Explainability Walkthrough}
\label{app:explain}

We provide the full trace summarized in Section~\ref{sec:explain}. During pre-training, we use two sample sentences from the sports domain. Initially, the clauses form random patterns using the features available in both samples, and neither sample satisfies the Boolean conjunction of literals, so the clause output is 0. This happens because the clause includes words like ``\textbf{player}'' and ``\textbf{professional}'', which are absent from Sample 2, and words like ``\textbf{defeat}'' and ``\textbf{opponent}'', which are absent from Sample 1. As learning progresses, Type I feedback (Reward, Penalty, and Inaction) adjusts the inclusion and exclusion of literals. This continues until the clause correctly identifies the semantic pattern for both samples (clause output 1), reaching the intended sports-cluster pattern [basketball AND winning AND team AND olympic].

In the fine-tuning phase with the vanilla TM, the clauses are no longer monotone and negated literals are reintroduced. Consider the labeled sports sample ``\textbf{After an intense game, the soccer team secured a stunning victory and the crowd erupted in celebration.}'' Its tokenized input is [intense, game, soccer, team, secured, stunning, victory, crowd, erupted, celebration]. The obvious sports-related features are [game, soccer, team, victory], which provide only a limited representation of the domain, so the clause may lack enough information to classify the sample as sports and instead relies mostly on negated features. After clustering, cluster-representation words are embedded into the sample, producing the enriched feature set [intense, game, soccer, team, secured, stunning, victory, crowd, erupted, celebration, basketball, winning, team, olympic]. The enriched input adds semantic information about sports, represented as [soccer, team, victory, basketball, winning, olympic] with only a few negated features, improving the model's semantic understanding of the sports class in a manner similar to BERT-based embeddings.

\end{document}